\documentclass{article}
\usepackage[utf8]{inputenc}

\title{RelNet: End-to-End Modeling of Entities \& Relations}
\author{
Trapit Bansal, 
Arvind Neelakantan,
Andrew McCallum \\
University of Massachusetts Amherst\\
\texttt{\{tbansal, arvind, mccallum\}@cs.umass.edu} \\
}
\date{May 2017}

\usepackage{graphicx}
\usepackage[final]{nips_2017}
\usepackage{amsmath}
\usepackage{hyperref}
\usepackage{cleveref}
\usepackage{xcolor}
\hypersetup{
	colorlinks,
	linkcolor={red!50!black},
	citecolor={blue!50!black},
	urlcolor={blue!80!black}
}

\begin{document}
\maketitle

\begin{abstract}
We introduce RelNet: a new model for relational reasoning.
RelNet is a memory augmented neural network which models entities as abstract memory slots and is equipped with an additional \textit{relational memory} which models relations between all memory pairs. 
The model thus builds an abstract knowledge graph on the entities and relations present in a document which can then be used to answer questions about the document. 
It is trained end-to-end: only supervision to the model is in the form of correct answers to the questions.
We test the model on the 20 bAbI question-answering tasks with 10k examples per task and find that it solves all the tasks with a mean error of 0.3\%, achieving 0\% error on 11 of the 20 tasks.
\end{abstract}

\section{Introduction}
Reasoning about entities and their relations is an important problem for achieving general artificial intelligence. Often such problems are formulated as reasoning over graph-structured representation of knowledge.
Knowledge graphs, for example, consist of entities and relations between them \citep{yago2,freebase,nell, auer2007dbpedia}. Representation learning \citep{riedel2013relation, bordes2013translating, wang2014knowledge, lin2015modeling} and reasoning \citep{das2016incorporating, neelakantan2015neural, neelakantan2016learning, miller2016key} with such structured representations is an important and active area of research.

Most previous work on knowledge representation and reasoning relies on a pipeline of natural language processing systems, often consisting of named entity extraction \citep{mccallum2003early}, entity resolution  and coreference \citep{dredze2010entity}, relationship extraction \citep{riedel2013relation}, and knowledge graph inference \citep{das2016chains}. While this cascaded approach of using NLP systems can be effective at reasoning with knowledge bases at scale, it also leads to a problem of compounding of the error from each component sub-system. The importance of each of these sub-component on a particular downstream application is also not clear. 

For the task of question-answering, we instead make an attempt at an end-to-end approach which directly models the entities and relations in the text as memory slots.
While incorporating existing knowledge (from curated knowledge bases) for the purpose of question-answering \citep{miller2016key, das2016incorporating, fader2014open} is an important area of research, we consider the simpler setting where all the information is contained within the text itself -- which is the approach taken by many recent memory based neural network models \citep{sukhbaatar2015end, henaff2017tracking, weston2015towards, munkhdalai2016reasoning}.

Recently, \cite{henaff2017tracking} proposed a dynamic memory based neural network for implicitly modeling the state of entities present in the text for question answering.
However, this model lacks any module for relational reasoning.
In response, we propose RelNet, which extends memory-augmented neural networks with a relational memory to reason about relationships between multiple entities present within the text. 
Our end-to-end method reads text, and writes to both memory slots and edges between them. Intuitively, the memory slots correspond to entities and the edges correspond to relationships between entities, each represented as a vector. The only supervision signal for our method comes from answering questions on the text.


We demonstrate the utility of the model through experiments on the bAbI tasks \citep{weston2015towards} and find that the model achieves smaller mean error across the tasks than the best previously published result \citep{henaff2017tracking} in the 10k examples regime and achieves 0\% error on 11 of the 20 tasks.

\section{RelNet Model}
\begin{figure}
    \centering
    \includegraphics[width=0.9\textwidth, height=2.1in]{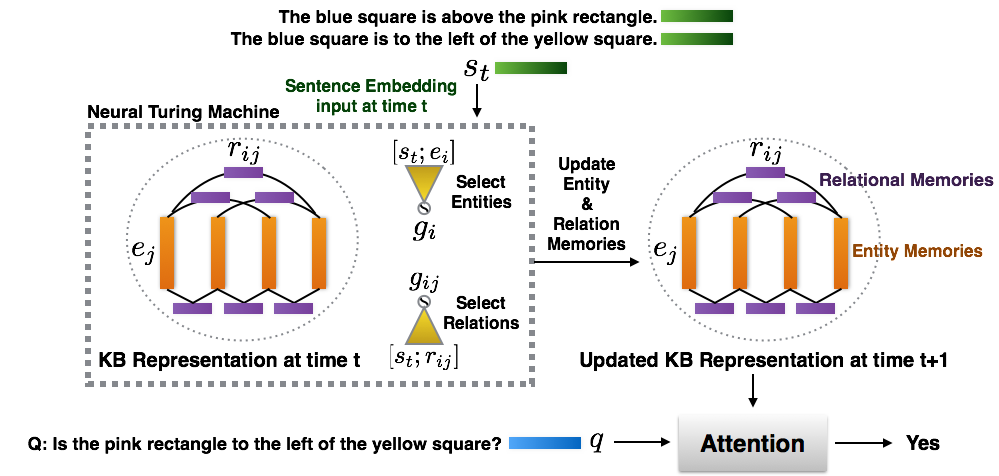}
    \caption{RelNet Model: The model represents the state of the world as a neural turing machine with relational memory. At each time step, the model reads the sentence into an encoding vector and updates both entity memories and all edges between them representing the relations.}
    \label{fig:relnet}
\end{figure}
We describe the RelNet model in this section. 
Figure \ref{fig:relnet} provides a high-level view of the model.
The model is sequential in nature, consisting of the following steps: read text, process it into a dynamic relational memory and then attention conditioned on the question generates the answer. We model the dynamic memory in a fashion similar to Recurrent Entity Networks \citep{henaff2017tracking} and then equip it with an additional relational memory.

There are three main components to the model: 1) input encoder 2) dynamic memory, and 3) output module. We will describe these three modules in details. The input encoder and output module implementations are similar to the Entity Network \citep{henaff2017tracking} and main novelty lies in the dynamic memory. We describe the operations executed by the network for a single example consisting of a document with $T$ sentences, where each sentence consists of a sequence of words represented with $K$-dimensional word embeddings $\{e_1, \ldots, e_N\}$, a question on the document represented as another sequence of words and an answer to the question. 

\paragraph{Input Encoder:} 
The input at each time point is a sentence from the document which can be encoded into a fixed vector representation using some encoding mechanism, such as a recurrent neural network. We use a simple encoder with a learned multiplicative mask \citep{henaff2017tracking, sukhbaatar2015end}:
$s_t = \sum_i m_i \odot e_i$. 

\paragraph{Dynamic Relational Memory}
This is the main component of an end-to-end reasoning pipeline, where we need to process the information contained in the text such that it can be used to reason about the entities, their properties and the relationships among them. The memory consists of two parts: entity memory and relational memory. The entity memory is organized as a key-value memory network \citep{miller2016key}, where the keys are global embeddings updated during training time but not during inference, and the value memory slot is a dynamic memory for each example (document, question) whose values are updated while reading the document. 
The memory thus consists of $D$ memory slots $\{m_1, \ldots, m_D\}$ (each is a vector of dimension $K$) and associated keys $\{k_1, \ldots, k_D\}$ (again vectors of dimension $K$). At time $t$, after reading the sentence $t$ into a vector representation $s_t$, a gating mechanism decides the set of memories to be updated ($<\cdot,\cdot>$ denotes inner product):
\begin{equation}
    g^m_i \leftarrow \sigma\left( <s_t,\ m_i+k_i> \right) \label{gate_ent}
\end{equation}
Intuitively the memory slots can be thought of as entities. Indeed, \cite{henaff2017tracking} found that if they tie the key vectors to entities in the text then the memories contain information about the state of those entities. The update in \eqref{gate_ent} essentially does a soft selection of memory slots based on cosine distance in the embedding space. Note that there can be multiple entites in a sentence hence a sigmoid operation is more suitable, and it is also more scalable \citep{henaff2017tracking}. 
After selecting the set of memories, there is an update step which stores information in the corresponding memory slots:
\begin{align}
    &\tilde{m}_j \leftarrow \mbox{PReLU}(Um_j + Vk_j + Ws_t) \nonumber\\
    &m_j \leftarrow m_j + g^m_i \odot \tilde{m}_j
\end{align}
where PReLU is a parametric Rectified linear unit \citep{he2015delving}, and $U$, $V$ and $W$ are $k\times k$ parameter matrices.

Now we augment the model with additional \textit{relational memory cells}. Intuitively, the entity memory allows modeling of entities and information about the entities in isolation. This can be insufficient in scenarios where a particular entity participates in may relations with other entities across the document. Thus, in order to succeed at relational reasoning the model needs to be able to compare each pair of the entity memories. The relational memories will allow modeling of these relations and provide an inherent inductive bias towards a more structured representation of the participating entities in the text, in the form a latent knowledge graph. The relational memories are $D^2$ memory slots $\{r_{ij}\}$ indexed by the entity memory slots $i,j \in D$.

The relational memories are updated as follows. First, a gating mechanism decides the set of active relational memories:
\begin{equation}
    g^r_{ij} \leftarrow g^m_i g^m_j \sigma( <s_t,\ r_{ij}>) \label{gate_rel}
\end{equation}
where $ g^m_i, g^m_j$ select the relational memory slot based on the active entity slots and the last sigmoid gate decides whether the corresponding relational memory needs to be updated based on the current input sentence. After selecting the set of active relational memory, we update the contents of the relational memory:
\begin{align}
    &\tilde{r}_{ij} \leftarrow PReLU(A r_{ij} + B s_t) \nonumber\\
    &r_{ij} \leftarrow r_{ij} + g^r_{ij} \odot \tilde{r}_{ij} \label{rel_update}
\end{align}
where again $A, B$ are $k\times k$ parameter matrices. Note that for updates \eqref{gate_rel}--\eqref{rel_update} we use a different encoding mask to obtain the sentence representation for relations.

Similar to \cite{henaff2017tracking}, we normalize the memories after each update step (that is after reading each sentence). This acts as a forget step and does not cause the memory to explode.

The full memory consists of the entity memory slots $\{h_j\}$ and the relational memory slots $\{r_{ij}\}$.

\paragraph{Output Module}
This is a standard attention module used in memory networks \citep{sukhbaatar2015end, henaff2017tracking}.
The question is encoded as a $K$ dimensional vector $q$ using the same encoding mechanism as the sentences (though with a separate learned mask). 
We first concatenate the relational memory vectors with the corresponding entity vectors, and project the resulting memory vector to $k$ dimension. Then attention on these projected memories, conditioned on the vector $q$, yields the final answer:
\begin{align*}
     & m^f_{ij} = C [m_i; m_j; r_{ij}] \\
     & p_{ij} = Softmax(<q, m^f_{ij}>) \\
     & u = \sum_{ij} p_{ij} m^f_{ij} \\
     & o = \mbox{PReLU}(q + Hu) \\
     & y = Z o
\end{align*}
where $y$ is the predicted answer, and $C, H, Z$ are parameter matrices.

\section{Related Work}
There is a long line of work in textual question-answering systems \citep{kwiatkowski2010inducing, berant2013semantic}. 
Recent successful approaches use memory based neural networks for question answering, for example \cite{weston2014memory,weston2015towards, xiong2016dynamic, munkhdalai2016reasoning, henaff2017tracking}.
Our model is also a memory network based model and is also related to the neural turing machine \citep{graves2014neural}. As described previously, the model is closely related to the Recurrent Entity Networks model \citep{henaff2017tracking} which describes an end-to-end approach to model entities in text but does not directly model relations. Other approaches to question answering use external knowledge, for instance external knowledge bases \citep{bordes2015large, miller2016key, das2017question, andreas2016learning, neelakantan2015neural} or external text like Wikipedia \citep{yang2015wikiqa, chen2017reading}.

Very recently, and in parallel to this work, a method for relational reasoning called relation networks \citep{santoro2017simple} was proposed. They demonstrated that simple neural network modules are not as effective at relational reasoning and their proposed module is similar to our model. However, relation network is not a memory-based model and there is no mechanism to read and write relevant information for each pair. Moreover, while their approach scales as the square of the number of sentences, our approach scales as the square of the number of memory slots used per QA pair. The output module in our model can be seen as a type of relation network.

Representation learning and reasoning over graph structured data is also relevant to this work. Graph based neural network models \citep{li2015gated, scarselli2009graph, kipf2016semi} have been proposed which take graph data as an input. The relational memory however does not rely on a specified graph structure and such models can potentially be used for multi-hop reasoning over the relational memory. \cite{johnson2016learning} proposed a method for learning a graphical representation of the text data for question answering, however the model requires explicit supervision for the graph at every step whereas RelNet does not require explicit supervision for the graph.

\begin{table}[t!]
    \centering
    \begin{tabular}{|l|c|c|}
\hline 
    \multicolumn{1}{|c|}{Task} & EntNet \cite{henaff2017tracking} & RelNet \\
\hline
1: 1 supporting fact & 0 & 0 \\
2: 2 supporting facts & 0.1 & 0.7 \\
3: 3 supporting facts & 4.1 & 3.4 \\
4: 2 argument relations & 0 & 0 \\
5: 3 argument relations & 0.3 & 0.6 \\
6: yes/no questions & 0.2 & 0 \\
7: counting & 0 & 0.1 \\
8: lists/sets & 0.5 & 0 \\
9: simple negation & 0.1 & 0 \\
10: indefinite knowledge & 0.6 & 0.1 \\
11: basic coreference & 0.3 & 0.1 \\
12: conjunction & 0 & 0 \\
13: compound coreference & 1.3 & 0 \\
14: time reasoning & 0 & 0.2 \\
15: basic deduction & 0 & 0 \\
16: basic induction & 0.2 & 0.1 \\
17: positional reasoning & 0.5 & 0 \\
18: size reasoning & 0.3 & 0.4 \\
19: path finding & 2.3 & 0 \\
20: agents motivation & 0 & 0 \\
\hline
\hline
Tasks with 0 \% error & 7 & \textbf{11} \\
\hline
Mean \% Error & 0.5 & \textbf{0.3} \\
\hline

\end{tabular}
    \caption{Mean \% Error on the 20 Babi tasks.}
    \label{tab:babi}
\end{table}

\section{Experiments}
We evaluate the model's performance on the bAbI tasks \citep{weston2015towards}, a collection of 20 question answering tasks which have become a benchmark for evaluating memory-augmented neural networks. We compare the performance with the Recurrent Entity Networks model (EntNet) \citep{henaff2017tracking}. Performance is measured in terms of mean percentage error on the tasks.

\textbf{Training Details:}
We used Adam and did a grid search for the learning rate in \{0.01, 0.005, 0.001\} and choose a fixed learning rate of 0.005 based on performance on the validation set, and clip the gradient norm at 2. 
We keep all other details similar to \cite{henaff2017tracking} for a fair comparison. embedding dimensions were fixed to be 100, models were trained for a maximum of 250 epochs with mini-batches size of 32 for all tasks except 3 for which the batch size was 16. The document sizes were limited to most recent 70 sentences for all tasks, except for task 3 for which it was limited to 130.
The RelNet models were run for 5 times with random seed on each task and the model with best validation performance was chosen as the final model. The baseline EntNet model was run for 10 times for each task \citep{henaff2017tracking}.

The results are shown in Table \ref{tab:babi}. The RelNet model achieves a mean error of \textbf{0.285\%} across tasks which is better than the results of the EntNet model \citep{henaff2017tracking}. The RelNet model is able to achieve 0\% test error on 11 of the tasks, whereas the EntNet model achieves 0\% error on 7 of the tasks.

\section{Conclusion}
We demonstrated an end-to-end trained neural network augmented with a structured memory representation which can reason about entities and relations for question answering. 
Future work will investigate the performance of these models on more real world datasets, interpreting what the models learn, and scaling these models to answer questions about entities and relations from reading massive text corpora.

\bibliographystyle{unsrtnat}
\bibliography{main}
\end{document}